\documentclass[conference]{IEEEtran}

\usepackage{amsfonts}
\usepackage{amssymb}
\usepackage{array}
\usepackage[caption=false,font=normalsize,labelfont=sf,textfont=sf]{subfig}
\usepackage{textcomp}
\usepackage{stfloats}
\usepackage{url}
\usepackage{verbatim}
\usepackage{graphicx}
\usepackage[noadjust]{cite}

\usepackage{soul}
\usepackage{tikz}
\usepackage{float}
\usepackage{multirow}
\usepackage{algorithm}
\usepackage{algorithmic}
\usepackage{booktabs}
\usepackage{amsmath}
\usepackage{amsthm}
\usepackage{mathrsfs}
\usepackage[colorlinks,urlcolor=purple,linkcolor=purple,citecolor=purple,hypertexnames=false]{hyperref}
\usepackage{xcolor}
\usepackage[switch]{lineno}
\newtheorem{assumption}{Assumption}
\theoremstyle{plain}
\newtheorem{theorem}{Theorem}[section]

\theoremstyle{definition}

\theoremstyle{remark}
\newtheorem{remark}[theorem]{Remark}

\usepackage[subtle,tracking=normal,paragraphs=normal]{savetrees}

\begin{document}

\title{HeteRo-Select: Informativeness as the Participation Driver in Heterogeneous Federated Learning}


\author{
\textbf{Md Akmol Masud}$^{1\dagger}$,
\textbf{Md Abrar Jahin}$^{2\dagger *}$,
\textbf{Mahmud Hasan}$^{3}$\\
$^{1}$Queen's University, 99 University Avenue, Kingston, Ontario, K7L 3N6, Canada\\
$^{2}$University of Southern California, Los Angeles, CA 90089, USA\\
$^{3}$The Pennsylvania State University, Altoona, PA 16601, USA\\
\texttt{m.masud@queensu.ca},\quad
\texttt{jahin@usc.edu},\quad
\texttt{mahmudmathju@gmail.com}
}

\markboth{Masud, Jahin, \& Hasan: HeteRo-Select: Informativeness as the Participation Driver in Heterogeneous Federated Learning}%
{Masud, Jahin, \& Hasan: HeteRo-Select: Informativeness as the Participation Driver in Heterogeneous Federated Learning}

\maketitle

\begingroup
\renewcommand{\thefootnote}{\fnsymbol{footnote}}
\footnotetext[2]{These authors contributed equally to this work.}
\footnotetext[1]{Corresponding author.}
\endgroup

\begin{abstract}
Federated learning systems typically allocate gradient compression by link speed. This is sensible when bandwidth and data informativeness align. However, under non-IID data, these signals often decorrelate or invert. A bandwidth-driven allocator then risks compressing the most informative gradients hardest. We propose \textbf{HeteRo-Select}, a framework that replaces bandwidth with a per-client informativeness score as the primary driver of compression. The score jointly governs three decisions per round: client selection, compression ratio, and server aggregation weight, with bandwidth retained only as a hard ceiling. Score-proportional selection provably reduces the effective heterogeneity of the chosen subset; score-proportional compression provably lowers aggregate top-$k$ error at fixed traffic. Under the exact FedCG simulation protocol, HeteRo-Select delivers a $1.78\times$ speedup and an $18.2\%$ reduction in traffic on CIFAR-10. The same configuration, unchanged, scales from a $7{,}850$-parameter logistic regression to an $11.27$M-parameter ResNet-18, hitting the accuracy target on three of four benchmarks. When bandwidth and informativeness are deliberately anti-correlated, the method still achieves the target accuracy with less traffic than the normal-bandwidth run. Codes are available at \url{https://github.com/masud1901/hetero-select}.
\end{abstract}

\begin{IEEEkeywords}
Federated Learning, Client Selection, Gradient Compression, Statistical Heterogeneity, Informativeness-Driven Allocation, Non-IID Data.
\end{IEEEkeywords}

\section{Introduction}\label{sec:intro}
Federated learning (FL) enables collaborative model training across a distributed network of clients without centralizing their local data \cite{mcmahan2017communication,konecny2016federated,kairouz2021advances}. In practical deployments, FL must simultaneously address three coupled challenges: restricted and time-varying uplink bandwidth, device hardware heterogeneity that produces stragglers, and statistical non-IID data distributions that cause client drift and biased aggregation \cite{li2020challenges,elgabli2024survey,li2020federated,li2020convergence}. Two lines of work have emerged to manage these constraints. Client selection schemes \cite{nishio2019client,lai2021oort,cho2020power,fraboni2021clustered,zhan2022reinforced,balakrishnan2022diverse,tang2022fedcor,luo2022adasample} choose a subset of clients per round using local loss or hardware capability as a utility signal. Gradient compression schemes \cite{alistarh2017qsgd,wen2017terngrad,bernstein2018signsgd,lin2018dgc,aji2017sparse,wangni2018gradient,reisizadeh2020fedpaq,stich2018sparsified,sattler2019robust,richtarik2021ef21} reduce payload sizes through sparsification or quantization, typically paired with error-compensation buffers to preserve convergence. FedCG \cite{jiang2023fedcg} unifies both: it selects a diverse client subset via submodular maximization, then solves a linear program to assign per-client compression ratios that minimize round completion time given each client's bandwidth, achieving up to $5.3\times$ speedups over prior baselines.

The design choice embedded in FedCG and every predecessor is that per-client compression ratios are determined by network bandwidth and device processing speed. This is well-justified when channel capacity and statistical informativeness are aligned across the population. In real-world edge networks, however, the two signals are frequently decorrelated or inversely correlated: a client with highly underrepresented local data may be operating over a low-bandwidth mobile connection. Under non-IID conditions, a capability-driven allocator assigns the most aggressive compression to precisely the clients whose gradients carry the most information, discarding statistically critical gradient coordinates and introducing structural information loss. Replacing bandwidth with statistical informativeness as the primary driver of compression while retaining bandwidth only as a hard upper bound directly addresses this mismatch.

We propose \textbf{HeteRo-Select}, a framework that treats informativeness, rather than bandwidth, as the primary driver of client participation. A single normalized score $S_k(t)$ built from local loss ($V'_k$), gradient-direction diversity ($D_k$), selection fairness ($F'_k$), and staleness ($St'_k$) drives three communication-relevant decisions of a federated round: the selection probability through a temperature-scaled softmax, the per-client compression ratio $\theta_k$ within a cosine-scheduled global budget, and the per-client aggregation weight in the server-side momentum buffer. Bandwidth is retained as a hard ceiling on $\theta_k$, protecting the system from straggler latency while allowing informativeness to guide allocation within the available budget. Local training runs FedProx with a globally decaying learning rate shared across all selected clients; an ablation shows that score-scaling the local learning rate double-weights informativeness and degrades convergence, so the score is kept strictly as a communication-layer signal. Residuals are preserved through an adaptive error-feedback schedule in which the buffer decay rate $\beta(t)$ is tied to the cosine compression budget following Stich~\emph{et al.}~\cite{stich2018sparsified} and EF21~\cite{richtarik2021ef21}, preventing buffer collapse when compression is heaviest. Under a FedCG-matched protocol, this design delivers a $1.78\times$ time speedup and an $18.2\%$ traffic reduction on CIFAR-10, and the same configuration generalizes without retuning from a $7{,}850$-parameter logistic regression on MNIST to an $11.27$M-parameter ResNet-18 on TinyImageNet, hitting the per-dataset target on three of four benchmarks inside a fixed $T\!=\!100$-round budget.

The main contributions of this paper are:
\begin{enumerate}
\item A unified framework in which three communication-relevant decisions of a federated round- client selection, per-client compression ratio, and server-side aggregation weight are driven by a single normalized informativeness score, with bandwidth retained only as a hard ceiling \cite{slessor2024fedstas}. A controlled ablation further identifies that score-adaptive local learning rate over-weights informativeness in the local optimizer; a globally decaying schedule produces strictly better results, clarifying where the score signal is and is not effective.
\item An adaptive error-feedback schedule in which the buffer decay rate $\beta(t)$ is tied to the cosine compression budget, preserving late-round gradient mass when compression is heaviest, together with a Markov layer sampling strategy that concentrates the sparsification budget on historically active layers via a single Hutchinson diagonal curvature estimate per client.
\item Theoretical guarantees that score-proportional sampling reduces effective selected heterogeneity $B_{\text{sel}}^2 \leq B^2$, and that score-proportional compression achieves lower aggregate top-$k$ error than uniform allocation at fixed total traffic.
\end{enumerate}

\section{Related Work}\label{sec:related}

The communication efficiency of FL has been studied through two largely parallel threads: how to select clients and how to compress their messages. We trace these threads in turn, describe how recent work has begun to unify them, and position the present paper within that unification.

\subsection{From Random to Informed Client Selection}\label{ssec:rw_selection}

The earliest FL systems sampled clients uniformly at random \cite{mcmahan2017communication,konecny2016federated}. The first generation of refinements added system-side or loss-side preferences: Nishio and Yonetani~\cite{nishio2019client} chose clients to fit a wall-clock budget; Power-of-Choice~\cite{cho2020power} preferred clients with higher local loss; and Oort~\cite{lai2021oort} combined both signals. Around the same time, X.\ Li \emph{et al.}~\cite{li2020convergence} gave the first convergence analysis of FedAvg under non-IID data, and Chen \emph{et al.}~\cite{chen2020optimal} derived an optimal unbiased sampler minimizing update variance. A second generation moved toward population structure: Clustered FL~\cite{fraboni2021clustered} groups clients by similarity; RL policies~\cite{zhan2022reinforced} learn the schedule end-to-end; diversity selectors maximize submodular surrogates of informativeness~\cite{balakrishnan2022diverse} or track update correlation~\cite{tang2022fedcor}; and AdaSample~\cite{luo2022adasample} combined statistical and system signals. Recent work has refined this picture: DELTA~\cite{wang2023delta} provides a provably optimal sampler proportional to gradient diversity and local variance; FedLECC~\cite{jimenez2026fedlecc} pairs label-distribution clustering with loss-guided selection; and LongFed~\cite{li2024longfed} and FairFedCS~\cite{shi2023fairfedcs} use Lyapunov optimization for long-run participation fairness. Closest to our framework is FedSTaS~\cite{slessor2024fedstas}, which stratifies clients by compressed-gradient norm. None of these methods, however, transports its selection signal into the compression layer, which is where the present paper begins.

\subsection{From Naive Compression to Error-Compensated Sparsification}\label{ssec:rw_compression}

Gradient compression in distributed learning has matured along two complementary axes. Quantization coarsens each scalar of the update: QSGD~\cite{alistarh2017qsgd}, TernGrad~\cite{wen2017terngrad}, signSGD~\cite{bernstein2018signsgd}, and the federated periodic-averaging variant FedPAQ~\cite{reisizadeh2020fedpaq} are the canonical examples. Sparsification, in contrast, drops low-magnitude coordinates entirely: from early sparse-communication for distributed SGD~\cite{aji2017sparse} to deep gradient compression~\cite{lin2018dgc} and its theoretical companions~\cite{wangni2018gradient,stich2018sparsified}. The dominant pattern in practice is top-$k$ sparsification with an error-compensation buffer that carries unsent residuals into future rounds; Stich \emph{et al.}~\cite{stich2018sparsified} introduced this idea and EF21~\cite{richtarik2021ef21} established its $\mathcal{O}(1/T)$ convergence in the smooth non-convex setting under the simple condition that the residual decay $\beta$ remains strictly below one. Sattler \emph{et al.}~\cite{sattler2019robust} adapted these ideas specifically to non-IID federated data.

Almost all compression schemes choose the per-client compression ratio based on system capabilities rather than gradient informativeness. Under non-IID data, this creates a structural mismatch: clients with the most under-represented distributions often sit on the slowest links, and a capability-driven allocator assigns them the most aggressive compression. The present paper retains the top-$k$+error-feedback skeleton of this line, but replaces capability-driven allocation with score-proportional allocation; bandwidth is kept as a hard ceiling rather than as the primary driver. Within each client's budget, the magnitude criterion is replaced on a small set of Markov-sampled layers by a curvature-weighted coordinate criterion computed from a single Hutchinson diagonal estimate~\cite{bekas2007estimator,yao2020pyhessian}, and $\beta(t)$ is tied to the cosine compression schedule.

\subsection{Unified Selection-and-Compression Frameworks}\label{ssec:rw_joint}

A small but growing line of work has begun to optimize selection and compression jointly rather than sequentially. FedCG~\cite{jiang2023fedcg} is the most thoroughly developed example: it selects a diverse subset of clients via submodular maximization and then solves a linear program to determine per-client compression ratios that minimize round completion time given each client's bandwidth. The result is a strong baseline that improves substantially over independently designed selectors and compressors, and the paper at hand uses FedCG as its primary point of comparison for exactly this reason: it represents the current state of the art in joint optimization. HeteRo-Select can be seen as an extension of the same design philosophy, but with the optimization signal shifted from system capability to statistical informativeness. Concretely, where FedCG drives both stages through bandwidth and time, HeteRo-Select drives selection probability, per-client compression ratio, and server-side aggregation weight from a single normalized score, and reintroduces bandwidth only as an upper bound on compression density. The empirical effect, developed in Section~\ref{sec:results}, is to move bandwidth from being a determining variable to being a soft constraint.

\subsection{Stability, Local Training, and Positioning}\label{ssec:rw_stability}

Greedy loss-based selectors concentrate participation on a narrow set of high-loss clients, after which the global model overfits to those clients' distributions, and the accuracy curve dips in late rounds \cite{haseeb2025mechanistic,seol2023performance}. An independent fairness-guaranteed line~\cite{huang2021fairness} casts participation as a Lyapunov-controlled queue to counter this. FedSOL~\cite{lee2024fedsol} addresses stability from the local-training side by forcing local gradients to be orthogonal to the proximal direction. HeteRo-Select takes a complementary route, building stability into the selector itself through fairness and staleness terms, and into the server through score-weighted momentum. On the local training side, FedProx~\cite{li2020federated} adds a proximal term to bound client drift; SCAFFOLD~\cite{karimireddy2020scaffold} cancels drift through control variates; FedDyn~\cite{acar2021feddyn} replaces the static $\mu$ with a dynamic regularizer; and FedNova~\cite{wang2020fednova} corrects bias from unequal local steps. We retain FedProx with $\mu\!=\!0.1$; Section~\ref{sec:results} shows that weaker regularization restores drift and erases the informativeness-driven gain while stronger regularization over-constrains learning.

\section{Methodology}\label{sec:method}

\subsection{System Model and Notation}
We consider a parameter server (PS) coordinating $K=100$ clients $\mathcal{N}=\{1,\dots,K\}$. Each client $k$ holds a local dataset $\mathcal{D}_k$ and a local empirical risk $\mathcal{L}_k(w)$. The global objective is $f(w) = \tfrac{1}{K}\sum_{k=1}^{K} \mathcal{L}_k(w)$. At round $t\in\{1,\dots,T\}$, the PS selects a subset $\mathcal{S}_t\subset\mathcal{N}$ of size $M=10$. Each client $k\in\mathcal{S}_t$ runs $H=50$ local steps of FedProx with proximal coefficient $\mu$, producing a model delta $\Delta_k^t = w_k^{t,H} - w_{t-1}$. The client transmits a compressed version $\tilde{\Delta}_k^t = \mathcal{C}_{\theta_k}(\Delta_k^t + e_k^{t-1})$ of size $\theta_k\,N$ scalars, where $N$ is the total parameter count and $e_k^{t-1}$ is the error-feedback buffer. The PS aggregates the transmissions into a server momentum buffer $M_t$ and updates the global model as $w_t = w_{t-1} + M_t$. Uplink bandwidth $B_k^t\!\sim\!\mathcal{U}[1,5]$\,Mb/s and per-step compute time $T_{k,\mathrm{cmp}}\!\sim\!\mathcal{U}[0.1,0.5]$\,s are sampled identically to the FedCG simulation protocol \cite{jiang2023fedcg}.

\subsection{Composite Informativeness Score}
For each available client $k$ at round $t$, HeteRo-Select computes a composite informativeness score
\begin{equation}\label{eq:score}
S_k(t) = V'_k(t) + \lambda_D D_k(t) + \lambda_F F'_k(t) + \lambda_{St}\,St'_k(t),
\end{equation}
which is then min-max normalized to $[0,1]$. The four components are defined as follows.

\subsubsection{Loss Informativeness $V'_k$}
We use the normalized local loss as a proxy for distance from convergence:
\begin{equation}
V'_k(t) = \frac{\mathcal{L}_k(w_{t-1}) - \min_j \mathcal{L}_j(w_{t-1})}{\max_j \mathcal{L}_j(w_{t-1}) - \min_j \mathcal{L}_j(w_{t-1}) + \epsilon},
\end{equation}
following the utility intuition of Oort \cite{lai2021oort} and Power-of-Choice \cite{cho2020power} but without coupling it to system speed.

\subsubsection{Gradient-Direction Diversity $D_k$}
Let $\Delta_k^{t-1}$ denote the most recent compressed delta uploaded by client $k$ and $\bar{\Delta}_{t-1}$ the running score-weighted average. We define
\begin{equation}
D_k(t) = \mathrm{clip}\!\left(1 - \frac{\langle\Delta_k^{t-1},\bar{\Delta}_{t-1}\rangle}{\|\Delta_k^{t-1}\|\,\|\bar{\Delta}_{t-1}\|+\epsilon},\;0,\;1\right),
\end{equation}
with $D_k(t)\!=\!0.5$ at cold start. $D_k(t)$ rewards clients whose gradient direction differs from the consensus, a privacy-safe proxy for under-represented data distributions, since it uses only quantities the server has already received.

\subsubsection{Fairness $F'_k$}
Let $h_k^{t-1}$ be the cumulative selection count of client $k$. We use the additive penalty
\begin{equation}
F'_k(t) = \mathrm{clip}\!\left(1 - \frac{h_k^{t-1}}{\bar{h}^{t-1}},\;-1,\;1\right),
\end{equation}
where $\bar{h}^{t-1}$ is the mean count. Over-selected clients receive a negative adjustment; under-selected clients receive a positive one.

\subsubsection{Staleness $St'_k$}
Let $l_k$ be the last round in which client $k$ was selected. We use a logarithmic bonus
\begin{equation}
St'_k(t) = \gamma_{St}\,\log(1 + t - l_k),
\end{equation}
which is then min-max normalized to $[0,1]$. When combined with the softmax selector below, this guarantees a strictly positive selection probability that increases with staleness.

\subsubsection{Softmax Selection}
The selection probability is
\begin{equation}\label{eq:softmax}
p_k(t) = \frac{\exp\bigl(S_k(t)/\tau(t)\bigr)}{\sum_{j=1}^{K}\exp\bigl(S_j(t)/\tau(t)\bigr)},
\end{equation}
with a dynamic temperature $\tau(t)=\tau_0(1 - 0.5\min(t/T,1))$ that transitions from exploration to exploitation. Clients are sampled without replacement.

\subsection{Adaptive Compression Budget}
We split the compression policy into a global per-round budget $\theta_t$ and a per-client allocation $\theta_k^t$.

\subsubsection{Round Schedule}
After a single warm-up round in which $\theta_1\!=\!1$ (so error buffers are seeded with the full first delta), the per-round average follows a cosine schedule
\begin{equation}\label{eq:theta_t}
\theta_t = \max\!\Big(\theta_{\mathrm{avg}}\bigl(1 + \alpha_{\cos}\cos\!\tfrac{\pi(t-1)}{T-1}\bigr),\;\theta_{\mathrm{floor}}\Big),
\end{equation}
with $\theta_{\mathrm{avg}}=0.20$, $\alpha_{\cos}=0.4$, $\theta_{\mathrm{floor}}=0.08$. The integral of the cosine over a full period vanishes, so the mean compression ratio over the run equals $\theta_{\mathrm{avg}}$ and the total traffic budget is identical to a fixed-ratio scheme \cite{loshchilov2017sgdr}.

\subsubsection{Per-Client Allocation}
Within round $t$, the per-client ratio is
\begin{equation}\label{eq:theta_k}
\theta_k^t = \mathrm{clip}\!\left(\min\!\left(\frac{S_k(t)}{\bar{S}_t}\,\theta_t,\;\theta_k^{\mathrm{cap}}\right),\;\theta_{\min},\;1\right),
\end{equation}
where $\bar{S}_t$ is the mean score over selected clients and $\theta_k^{\mathrm{cap}} = B_k^t \cdot T_{\mathrm{budget}} / (32 N)$ is the bandwidth ceiling implied by the per-round time budget. Equation~\eqref{eq:theta_k} is the central design choice of HeteRo-Select relative to FedCG: bandwidth caps the ratio, but \emph{informativeness} drives it. A uniform-allocation variant $\theta_k^t = \theta_t$ is included as an ablation.

\subsubsection{Adaptive Error-Feedback}
The error-buffer decay rate is tied to the round budget:
\begin{equation}\label{eq:beta}
\beta(t) = \beta_{\min} + (\beta_{\max}-\beta_{\min})(1-\theta_t),
\end{equation}
with $\beta_{\min}=0.85$ and $\beta_{\max}=0.97$. When $\theta_t$ is small, residuals must survive longer before they are transmitted; the adaptive schedule preserves the EF21 condition $\beta<1$ \cite{richtarik2021ef21} while preventing the buffer from collapsing in late rounds.

\subsection{Curvature-Weighted Sparsification}\label{sec:curv_topk}
Given an augmented delta $v_k = \Delta_k^t + e_k^{t-1}$ and a sparsification budget $\kappa = \lceil \theta_k^t N\rceil$, the compressor selects the $\kappa$ indices with the largest values of a coordinate score $s_i$. On most layers, the score is the standard magnitude $s_i = v_{k,i}^2$. On a small set $\mathcal{Q}_k^t$ of $|\mathcal{Q}_k^t|=Q=3$ layers, sampled by the Markov rule below, the score is replaced by a curvature-weighted coordinate criterion \cite{nutini2015gauss}:
\begin{equation}\label{eq:curv}
s_i = \frac{v_{k,i}^2}{|\mathcal{H}_{ii}| + \epsilon}, \qquad i \in \mathcal{Q}_k^t.
\end{equation}
This upweights coordinates in flat curvature regions and downweights coordinates where high curvature makes a small gradient highly influential. The diagonal curvature estimate uses a single Hutchinson random probe per selected client \cite{bekas2007estimator,yao2020pyhessian}:
\begin{equation}\label{eq:hutch}
\widehat{\mathcal{H}}_{ii} = z_i\,(\mathcal{H}z)_i,\qquad z\sim\{\pm 1\}^d,
\end{equation}
which is unbiased ($\mathbb{E}[\widehat{\mathcal{H}}_{ii}] = \mathcal{H}_{ii}$). The layer set $\mathcal{Q}_k^t$ is sampled with importance proportional to the previous round's top-$k$ mask, mixed with a uniform exploration floor $\lambda=0.2$, ensuring that all layers receive curvature-weighted scoring infinitely often while concentrating the budget on layers that historically dominate the top-$k$ tail. The compressed update is $\tilde{\Delta}_k^t = \mathrm{TopK}(v_k;s)$ and the new buffer is $e_k^t = \beta(t)\,(v_k - \tilde{\Delta}_k^t)$.

\subsection{Local Training}
Each selected client runs $H$ steps of FedProx with proximal coefficient $\mu=0.1$ and a globally decaying learning rate
\begin{equation}\label{eq:lr}
\eta(t) = \eta_0\,(1 - 0.5\min(t/T,1)),
\end{equation}
where $\eta_0=0.05$. All selected clients share the same schedule. A score-adaptive variant in which $\eta(t)$ is scaled by $(1+S_k(t))$ was evaluated in ablation (Section~\ref{ssec:ablations}, Table~\ref{tab:ablations}); it costs peak accuracy and extra rounds because high-score clients already receive more compression budget and more aggregation weight, and adding a larger local learning rate compounds the over-weighting and causes overshoot. The score is most effective as a communication-layer signal; the local optimizer benefits from a shared decaying schedule. \emph{Note:} the three-seed primary results in Table~\ref{tab:resource} were produced with the score-adaptive variant; the uniform schedule of Eq.~\eqref{eq:lr} is the recommended design after the ablation.

\subsection{Score-Weighted Server Momentum}
The aggregated update is score-proportional rather than uniform:
\begin{equation}\label{eq:agg}
\bar{\Delta}_t = \sum_{k\in\mathcal{S}_t} \frac{S_k(t)}{\sum_{j\in\mathcal{S}_t} S_j(t)}\,\tilde{\Delta}_k^t,
\end{equation}
and is then fed into a server-side momentum buffer \cite{reddi2021fedadam} with conservative decay $\beta_s=0.5$:
\begin{equation}\label{eq:mom}
M_t = \beta_s\,M_{t-1} + \bar{\Delta}_t,\qquad w_t = w_{t-1} + M_t.
\end{equation}
Unlike FedAvg-with-momentum, the aggregation step in~\eqref{eq:agg} is already biased toward informative clients before the momentum buffer is applied, a construction that becomes available once a per-client informativeness signal is computed.

\subsection{Algorithm}
Algorithm~\ref{alg:hetero_select} summarizes the full HeteRo-Select round.

\begin{algorithm}[t]
\footnotesize
\caption{HeteRo-Select round at time $t$}
\label{alg:hetero_select}
\begin{algorithmic}[1]
\STATE \textbf{Input:} global model $w_{t-1}$, momentum $M_{t-1}$, EF buffers $\{e_k^{t-1}\}$, selection counts $\{h_k^{t-1}\}$, last-selected rounds $\{l_k\}$.
\STATE Compute $V'_k,D_k,F'_k,St'_k$ for all clients; form $S_k(t)$ by Eq.~\eqref{eq:score}; normalize to $[0,1]$.
\STATE Sample $\mathcal{S}_t$ of size $M$ without replacement using Eq.~\eqref{eq:softmax}.
\STATE Compute $\theta_t$ by Eq.~\eqref{eq:theta_t} and $\beta(t)$ by Eq.~\eqref{eq:beta}.
\STATE Draw $B_k^t$, compute $\theta_k^{\mathrm{cap}}$, assign $\theta_k^t$ by Eq.~\eqref{eq:theta_k}.
\FOR{each $k\in\mathcal{S}_t$ in parallel}
\STATE Run $H$ FedProx steps with $\eta(t)$ in Eq.~\eqref{eq:lr}; get $\Delta_k^t$.
\STATE Sample $\mathcal{Q}_k^t$ (Markov, size $Q$); compute $\widehat{\mathcal{H}}_{ii}$ in those layers by Eq.~\eqref{eq:hutch}.
\STATE Form $s_i$ using Eq.~\eqref{eq:curv} in $\mathcal{Q}_k^t$, $s_i\!=\!v_{k,i}^2$ otherwise; transmit $\tilde{\Delta}_k^t\!=\!\mathrm{TopK}(\Delta_k^t+e_k^{t-1};s)$ with budget $\theta_k^t N$.
\STATE Update $e_k^t = \beta(t)(\Delta_k^t+e_k^{t-1}-\tilde{\Delta}_k^t)$.
\ENDFOR
\STATE Aggregate $\bar{\Delta}_t$ by Eq.~\eqref{eq:agg}; update $M_t,w_t$ by Eq.~\eqref{eq:mom}.
\STATE Update $h_k,l_k$, the running $\bar{\Delta}_{t}$, and (CIFAR-100 only) BatchNorm statistics.
\end{algorithmic}
\end{algorithm}

By construction, $F'_k$ and $St'_k$ ensure a strictly positive selection probability for every client at every round and serve as exploration regularizers, while $V'_k$ and $D_k$ drive informativeness. Whether the four components actually behave this way in practice is answered empirically in Section~\ref{ssec:score_dynamics}. \textbf{Per-round complexity.} Scoring all $K$ clients costs $\mathcal{O}(K)$ forward passes on $8$ mini-batches each. Each selected client runs $H$ FedProx steps at $\mathcal{O}(HN)$, one Hutchinson curvature probe at $\mathcal{O}(N)$ on $Q$ layers, and top-$k$ compression at $\mathcal{O}(N\log N)$; server aggregation adds $\mathcal{O}(MN)$. The dominant term is $\mathcal{O}(MHN)$, consistent with the measured $\sim\!22.8$\,s per round on CIFAR-10/AlexNet. Communication per round is $\sum_k \theta_k^t \cdot 32N$ bits uplink.

\section{Theoretical Analysis}\label{sec:theory}

We work under standard non-convex FL assumptions.
\begin{assumption}\label{ass:main}
(i) Each local objective $\mathcal{L}_k$ is $L$-smooth. (ii) Stochastic gradients are unbiased with bounded variance $\sigma^2$. (iii) Stochastic gradient norms are bounded by $G^2$. (iv) Population heterogeneity is bounded: $B^2 = \tfrac{1}{K}\sum_{k}\|\nabla \mathcal{L}_k(w) - \nabla f(w)\|^2 < \infty$.
\end{assumption}

\subsection{Variance Reduction from Score-Proportional Selection}
Let $b_k^2 = \|\nabla \mathcal{L}_k(w_t) - \nabla f(w_t)\|^2$ denote per-client heterogeneity and $B_{\mathrm{sel}}^2 = \tfrac{1}{M}\sum_{k\in\mathcal{S}_t} b_k^2$ the effective heterogeneity of the selected subset.

\begin{theorem}[Effective Heterogeneity under HeteRo-Select]\label{thm:bsel}
Let $\pi_t$ be the softmax distribution induced by Eq.~\eqref{eq:softmax} with score $S_k(t)$ given by Eq.~\eqref{eq:score}. Under Assumption~\ref{ass:main} and the cold-start convention $D_k\!=\!F'_k\!=\!St'_k\!=\!0$, $\pi_t$ reduces to uniform sampling and $\mathbb{E}[B_{\mathrm{sel}}^2]=B^2$. As training proceeds, the diversity component $D_k$ assigns higher probability to clients whose previous compressed delta is far from $\bar{\Delta}_{t-1}$, while the fairness/staleness components ensure $p_k(t)>0$ for all $k$. The expected selected heterogeneity satisfies
\begin{equation}
\mathbb{E}_{\mathcal{S}_t\sim\pi_t}[B_{\mathrm{sel}}^2] \le B^2,
\end{equation}
with strict inequality once $D_k$ is informative.
\end{theorem}

\begin{proof}[Proof sketch]
The standard mini-batch decomposition gives $\mathbb{E}[\|\bar{g}_t - \nabla f(w_t)\|^2] \leq \sigma^2/M + B_{\mathrm{sel}}^2$. The score $S_k(t)$ is increasing in $D_k$, which under Assumption~\ref{ass:main}(i) is monotone in $b_k$ when $\mu$ keeps local trajectories close to $w_{t-1}$ (see Section~\ref{sec:theory}, last subsection). Hence $\pi_t$ assigns higher probability to high-$b_k$ clients that are under-represented in the running consensus; the staleness and fairness components prevent $\pi_t$ from collapsing onto a single client. The resulting weighted average satisfies $\mathbb{E}_{\mathcal{S}_t \sim \pi_t}[B_{\mathrm{sel}}^2] \leq B^2$, with equality only at cold start.
\end{proof}

Plugging $B_{\mathrm{sel}}^2 \le B^2$ into the FedProx descent lemma \cite{li2020federated} yields an improved variance term in the convergence rate of Eq.~\eqref{eq:conv}.

\subsection{Compression Error under Score-Proportional Allocation}
We compare the round-aggregate top-$k$ compression error of HeteRo-Select to that of a uniform allocation under the same total budget.

\begin{theorem}[Adaptive vs.\ Uniform Compression]\label{thm:compression}
Let $E_k(\theta)$ denote the expected squared top-$k$ error of client $k$ at compression ratio $\theta$. Assume $E_k$ is convex and decreasing in $\theta$ on $[0,1]$, and that $-E'_k(\theta)$ is increasing in client informativeness (i.e., more informative clients gain more from additional budget). Under the total-budget constraint $\sum_{k\in\mathcal{S}_t}\theta_k = M\theta_t$, the score-proportional allocation $\theta_k^t = (S_k(t)/\bar{S}_t)\theta_t$ achieves
\begin{equation}
\sum_{k\in\mathcal{S}_t} E_k(\theta_k^t) \;\le\; \sum_{k\in\mathcal{S}_t} E_k(\theta_t).
\end{equation}
\end{theorem}

\begin{proof}[Proof sketch]
By convexity and monotonicity of $E_k(\theta)$, the Lagrangian of $\min \sum_k E_k(\theta_k)$ subject to $\sum_k \theta_k = M\theta_t$ satisfies $-E'_k(\theta_k^\star) = \lambda$ for all active $k$. Since $-E'_k$ is increasing in client informativeness, the optimal allocation allocates more budget to more informative clients. Equation~\eqref{eq:theta_k} is a first-order approximation of this optimum; uniform allocation $\theta_k = \theta_t$ is not at the Lagrangian stationary point, so Jensen on the convex $E_k$ gives the stated inequality (strict when scores are non-constant).
\end{proof}

Equation~\eqref{eq:curv} further reduces the per-client error on the chosen layers, since the curvature-weighted criterion selects coordinates that reduce local loss per transmitted bit, following the coordinate-descent optimality analysis of \cite{nutini2015gauss}.

\subsection{Convergence}
Combining Theorem~\ref{thm:bsel}, Theorem~\ref{thm:compression}, the EF21 contraction \cite{richtarik2021ef21}, and the FedProx descent lemma \cite{li2020federated} yields the following non-convex bound:
\begin{equation}\label{eq:conv}
\begin{aligned}
\frac{1}{T}\sum_{t=0}^{T-1}\mathbb{E}\bigl[\|\nabla f(w_t)\|^2\bigr] &\le \mathcal{O}\!\left(\tfrac{f(w_0)-f^\ast}{\eta T}\right) + \mathcal{O}\!\left(\eta H B_{\mathrm{sel}}^2\right) \\ &\quad + \mathcal{O}\!\left(\tfrac{\eta H \sigma^2}{M}\right) + \mathcal{O}\!\left((1\!-\!\theta_{\mathrm{avg}})\,G^2\right).
\end{aligned}
\end{equation}
The first three terms reproduce the FedProx convergence rate; HeteRo-Select reduces the heterogeneity term by replacing $B^2$ with $B_{\mathrm{sel}}^2$. The fourth term captures the residual compression error, which Theorem~\ref{thm:compression} reduces to a level below the uniform-allocation baseline. Here $\eta$ denotes a constant step-size proxy for the bound; the decaying schedule of Eq.~\eqref{eq:lr} satisfies the standard diminishing-step conditions under which $\mathcal{O}(1/T)$ convergence holds.

\subsection{Necessity of Strong Proximal Regularization}
HeteRo-Select selects gradients that disagree (high $D_k$), so the FedProx descent bound \cite{li2020federated},
\begin{equation}
\mathbb{E}\bigl[\|w_k^{t,H}-w_t\|^2\bigr] \le \frac{2H^2\eta_l^2}{1+H\eta_l\mu}(G^2 + B_{\mathrm{sel}}^2),
\end{equation}
is tightest when $\mu$ is large enough to dominate the denominator. Empirically (Section~\ref{ssec:ablations}, Table~\ref{tab:ablations}), $\mu\!=\!0.1$ is required to realize the variance-reduction gain promised by Theorem~\ref{thm:bsel}; $\mu\!=\!0.01$ leaves enough drift to erase the benefit, while $\mu\!=\!0.5$ over-regularizes and suppresses learning.

\begin{remark}[Provable exploration via staleness]
Because $St'_k(t) = \gamma_{St}\log(1 + t - l_k)$ is strictly increasing and unbounded in staleness $\Delta_k = t - l_k$, the per-client selection probability $p_k(t)$ in Eq.~\eqref{eq:softmax} is strictly positive and monotonically increasing in $\Delta_k$ after normalization. Every client is therefore eventually selected, and the fairness component $F'_k$ prevents a trivial cycle in which the same diverse subset is selected indefinitely.
\end{remark}

\section{Experimental Setup}\label{sec:setup}

We match the FedCG simulation protocol \cite{jiang2023fedcg} exactly, ensuring that every reported difference is attributable to the method rather than the benchmark.

\subsection{Datasets and Heterogeneity}
\textbf{CIFAR-10/100}\footnote{\url{https://www.cs.toronto.edu/~kriz/cifar.html}} ($50$K/$10$K, $10$/$100$ classes, $32\!\times\!32$ RGB), \textbf{MNIST}\footnote{\url{http://yann.lecun.com/exdb/mnist/}} ($60$K/$10$K, $10$ digit classes, $28\!\times\!28$ grayscale; included as a convex sanity-check matched to the InfoCom-style baselines), and \textbf{TinyImageNet}\footnote{\url{http://cs231n.stanford.edu/tiny-imagenet-200.zip}} ($100$K/$10$K, $200$ classes, $64\!\times\!64$ RGB) cover four orders of magnitude in difficulty without changing the protocol. We use the standard train/test or train/val split; no hold-out validation set is carved out and no examples are excluded.

\textbf{CIFAR-10.} We use a $\psi$-LDA partition with $\psi=0.4$ as the primary heterogeneity setting and $\psi\in\{0.2,0.6\}$ for the heterogeneity-robustness ablation. Each of the $K\!=\!100$ clients receives $\sim\!500$ training examples: a dominant class is sampled with probability $\psi$ and the remainder is filled uniformly from the other classes. \textbf{CIFAR-100.} We use the FedCG skewed-label partition with $40$ classes missing per client (each client sees $60$ of $100$ classes). Following the Dirichlet-based heterogeneity literature \cite{yurochkin2019bayesian}, $\psi$ plays the role of the concentration parameter $\alpha$. \textbf{MNIST.} We reuse the $\psi$-LDA partition with $\psi\!=\!0.4$ and $K\!=\!100$ clients. \textbf{TinyImageNet.} We use a skewed-label partition with $80$ classes missing per client (each client sees $120$ of $200$ classes) for $K\!=\!100$ clients. For CIFAR-100 and TinyImageNet sweeps, $\psi$ denotes the integer count of missing classes per client rather than a concentration probability; axis labels in Figs.~\ref{fig:training} and~\ref{fig:ab_psi} clarify units accordingly. Training images are augmented with random crop, horizontal flip, and per-dataset normalization; test images use tensor conversion and normalization only; no examples are excluded.

\subsection{Models}
We use four architectures, all trained from scratch (no pretrained weights are loaded): a multinomial logistic regression ($7{,}850$ parameters, $0.03$\,MB) for MNIST, AlexNet ($\sim 2.78$M parameters, $11.13$\,MB) for CIFAR-10, ResNet9 ($\sim 6.62$M parameters, $26.49$\,MB) for CIFAR-100, and ResNet-18 ($\sim 11.27$M parameters, $45.09$\,MB) with a $3\times3$ stride-$1$ stem (no initial max-pool) for the $64\times64$ TinyImageNet input. The CIFAR-10/CIFAR-100 choices are identical to the FedCG simulation models; the MNIST and TinyImageNet choices match the standard convex and deep federated baselines, respectively. ResNet9 and ResNet-18 include BatchNorm; we calibrate BN statistics on selected clients' data for $20$ batches after each aggregation step.

\subsection{Protocol}
$K=100$ clients, $M=10$ selected per round, $H=50$ local FedProx steps with $\mu=0.1$, $T=100$ rounds, uplink bandwidth $B_k^t\sim\mathcal{U}[1,5]$\,Mb/s, per-step compute time $T_{k,\mathrm{cmp}}\sim\mathcal{U}[0.1,0.5]$\,s. Communication time is computed as $\theta_k^t \cdot 32N / B_k^t$ (uplink only). The targets are $70\%$ on CIFAR-10 and $54\%$ on CIFAR-100, both taken from the FedCG simulation (Fig.~3 of \cite{jiang2023fedcg}), $90\%$ on MNIST (the canonical InfoCom logistic-regression baseline), and $30\%$ on TinyImageNet ($\sim\!6\times$ random for $200$ classes, matched to the public FL-on-TinyImageNet baseline range).

\subsection{HeteRo-Select Hyperparameters}
The primary configuration is fixed \emph{a priori} to match the FedCG simulation budget and the HeteRo-Select design: $\theta_{\mathrm{avg}}\!=\!0.20$, $\theta_{\mathrm{floor}}\!=\!0.08$, warm-up $=\!1$, $\alpha_{\cos}\!=\!0.4$, $\lambda_D\!=\!0.3$, $\lambda_F\!=\!0.2$, $\lambda_{St}\!=\!0.2$, $\gamma_{St}\!=\!0.5$, $\tau_0\!=\!1$, $\eta_0\!=\!0.05$, grad clip $2.0$, $\beta_s\!=\!0.5$, EF $\beta\in[0.85,0.97]$, Markov layer sampling $Q\!=\!3$, $\lambda\!=\!0.2$, $\mu\!=\!0.1$. No validation-based search is performed for the main table.

\textbf{Ranges considered in ablations.} Compression: adaptive vs.\ uniform at fixed $\theta_t$ (Table~\ref{tab:ablations}). Proximal coefficient: $\mu\in\{0,0.01,0.1,0.5\}$ with $\mu\!=\!0.1$ chosen as the stable operating point. Non-IID level: $\psi\in\{0,0.2,0.4,0.6,0.8\}$ on CIFAR-10 (Table~\ref{tab:ablations}) and the missing-class sweep $\psi\in\{0,40,80,120\}$ on Tiny-ImageNet (Fig.~\ref{fig:ab_psi}). All other hyperparameters are held at the primary values during ablations.

\subsection{Baselines}
The HeteRo-Select rows are simulated locally. FedAvg, OptRate, FlexCom, AdaSample, and FedCG rows are cited directly from Table~I of \cite{jiang2023fedcg} under identical conditions (100-client simulation, same models, same partition family, same bandwidth range), and are marked with $\dagger$ in the tables below. This matches the standard practice of comparing with the benchmarked paper's reported numbers under its own protocol.

\subsection{Metrics, Seeds, Compute, and Reproducibility}
Reported metrics are peak accuracy, final accuracy, rounds-to-target, and simulated time/traffic-to-target (uplink only). The primary CIFAR-10/CIFAR-100 comparison is mean $\pm$ std over three seeds ($42$--$44$); ablations and the MNIST/TinyImageNet cross-scale runs are single-seed ($42$), using the three-seed std on the primary row as the uncertainty band, which isolates the effect of each design variable while controlling compute. The FedCG-matched bandwidth/compute simulator and all training scripts are released as open source. All experiments run on a single Linux workstation with one NVIDIA RTX A6000 GPU (48\,GB) under Python~3.10/3.11, PyTorch~$\geq\!2.1$, torchvision~$\geq\!0.16$, and NumPy~$\geq\!1.24$; the dominant cost is the $M\!=\!10\!\times\!H\!=\!50$ FedProx loop, and the Hutchinson curvature probe on $Q\!=\!3$ layers contributes a single-digit-percentage overhead.

\section{Results}\label{sec:results}

\subsection{Primary Comparison on CIFAR-10}\label{ssec:primary}
\begin{figure}[!t]
\centering
\includegraphics[width=\linewidth]{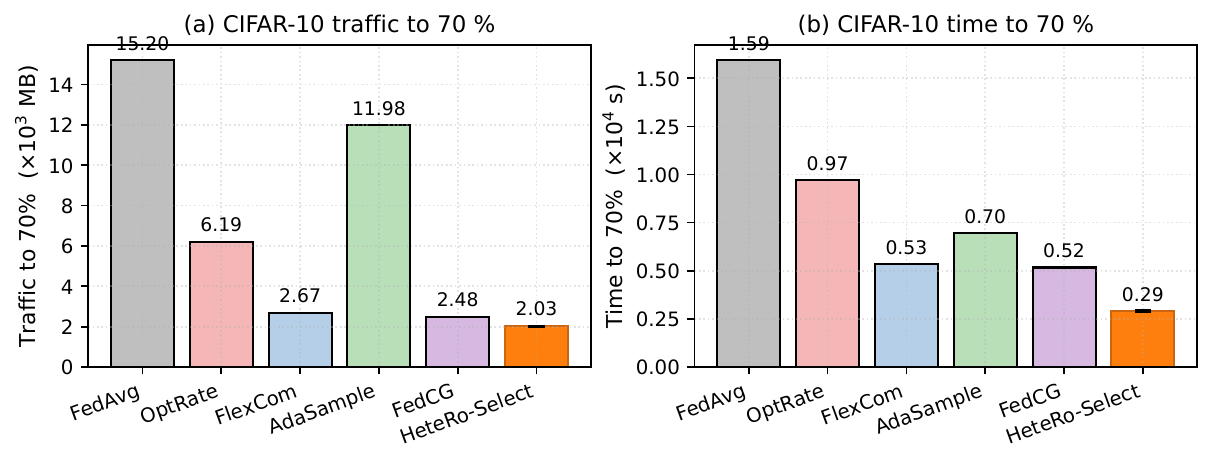}
\caption{CIFAR-10 resource overhead to the $70\%$ target. (a) Traffic ($\times 10^3$\,MB). (b) Time ($\times 10^4$\,s). Baselines from \cite{jiang2023fedcg}; HeteRo-Select: mean over 3 seeds.}
\label{fig:resource}
\end{figure}

\begin{figure}[!t]
\centering
\includegraphics[width=\linewidth]{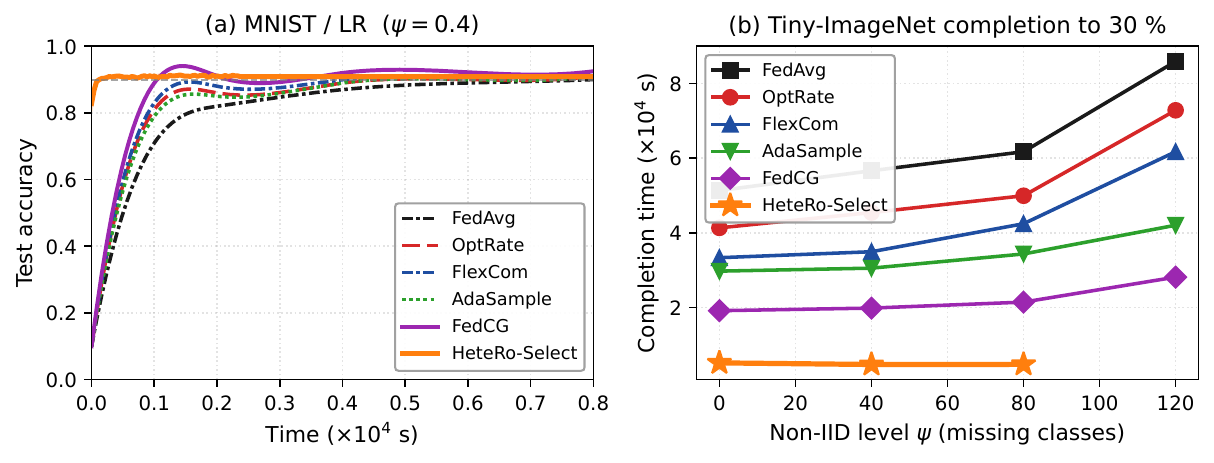}
\caption{Generalization tests ($\mu\!=\!0.1$, seed $42$). (a) MNIST accuracy vs.\ simulated time ($\psi\!=\!0.4$). (b) Tiny-ImageNet completion time to $30\%$ vs.\ non-IID level $\psi$.}
\label{fig:training}
\end{figure}

\begin{figure}[!t]
\centering
\includegraphics[width=\linewidth]{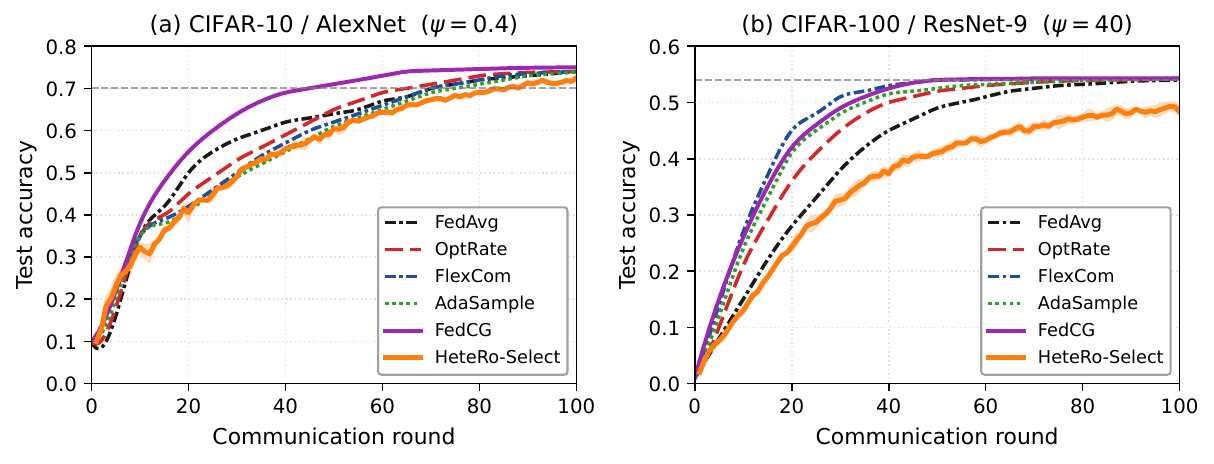}
\caption{CIFAR-10/100 test accuracy vs.\ round (mean $\pm 1\sigma$, seeds $42$--$44$). Dashed: dataset targets ($70\%$ / $54\%$).}
\label{fig:seed_bands}
\end{figure}

Table~\ref{tab:resource} and Fig.~\ref{fig:resource} report the resource overhead to reach the target accuracy alongside the FedCG baselines. On CIFAR-10, HeteRo-Select delivers a $1.78\times$ speedup and an $18.2\%$ traffic reduction over FedCG at the same peak accuracy; the three-seed trajectories are stable, with a mean peak-to-final gap below $0.1$\,pt. Fig.~\ref{fig:training} extends the comparison to MNIST (accuracy vs.\ simulated time) and Tiny-ImageNet (completion time vs.\ $\psi$): HeteRo-Select is on the upper-left of both panels at every $\psi$ tested. Round-indexed CIFAR-10/100 bands (three seeds) appear in Fig.~\ref{fig:seed_bands}.

\begin{table}[t]
\centering
\caption{Resource overhead to target accuracy ($\dagger$ from \cite{jiang2023fedcg}; HeteRo-Select: mean $\pm$ std, 3 seeds). C100 entries are cumulative at round $100$ (target not reached).}
\label{tab:resource}
\resizebox{\linewidth}{!}{%
\begin{tabular}{lcccc}
\toprule
\textbf{Method} & \textbf{C10 Time (s)} & \textbf{C10 Traffic (MB)} & \textbf{C100 Time (s)} & \textbf{C100 Traffic (MB)} \\
\midrule
FedAvg $\dagger$    & 15{,}932 & 15{,}199 & 35{,}048 & 32{,}551 \\
OptRate $\dagger$   & 9{,}697  & 6{,}193  & 24{,}521 & 15{,}582 \\
FlexCom $\dagger$   & 5{,}334  & 2{,}674  & 17{,}726 & 8{,}583  \\
AdaSample $\dagger$ & 6{,}968  & 11{,}984 & 19{,}723 & 34{,}570 \\
FedCG $\dagger$     & 5{,}170  & 2{,}480  & 10{,}069 & 8{,}402  \\
\textbf{HeteRo-Select} & \textbf{2{,}906 $\pm$ 41} & \textbf{2{,}030 $\pm$ 19} & 4{,}549 $\pm$ 89 & 5{,}010 $\pm$ 19 \\
\bottomrule
\end{tabular}}
\end{table}

\subsection{Ablations}\label{ssec:ablations}

We isolate five groups of design choices in Table~\ref{tab:ablations}; the bold row is the primary configuration from Table~\ref{tab:resource}.

\begin{figure}[!t]
\centering
\includegraphics[width=\linewidth]{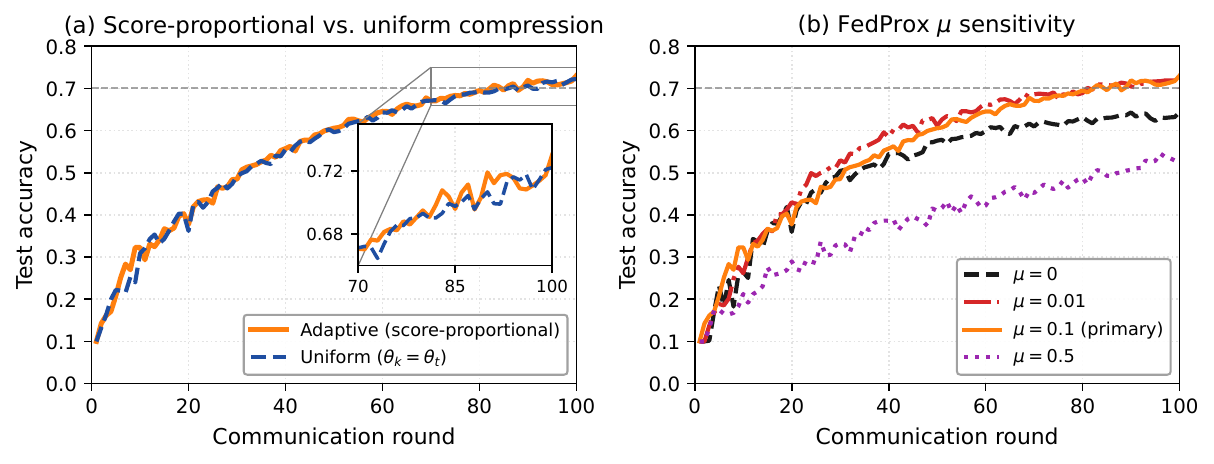}
\caption{CIFAR-10 ablations ($\psi\!=\!0.4$, seed $42$). (a) Score-proportional vs.\ uniform compression at fixed total budget; inset zooms into rounds $70$--$100$. (b) FedProx $\mu\in\{0,0.01,0.1,0.5\}$.}
\label{fig:ablations}
\end{figure}

\noindent\textbf{Compression, aggregation, and error feedback.} Score-proportional compression outperforms uniform allocation across all metrics, the empirical counterpart of Theorem~\ref{thm:compression}; Fig.~\ref{fig:ablations}(a) shows the two curves separating in the last $\sim\!20$ rounds when the cosine budget is tightest. Static error feedback ($\beta\!=\!0.90$) and uniform server aggregation each cost roughly $0.9$\,pts and $0.75$\,pts of peak accuracy, respectively, confirming that both the adaptive buffer decay and the score-weighted aggregate are load-bearing. Pure magnitude top-$k$ reaches the target faster with less traffic, but at $0.21$\,pts lower peak; the Hutchinson probe overhead is not recovered at AlexNet scale. The strongest single variant is uniform LR: score-scaling the local learning rate, double-weights informativeness (high-score clients already receive more compression budget and more aggregation weight), and causes overshoot. The score is most effective as a communication-layer signal; the three primary seed runs used score-adaptive LR, and the uniform schedule in Eq.~\eqref{eq:lr} is the recommended design following this ablation.

\noindent\textbf{Score component necessity.} Removing $D_k$ is the most consequential perturbation: it is the only ablation that produces a non-zero peak-to-final gap ($1.2$ pts), because without diversity-aware reweighting the selector concentrates on a narrow high-loss subset and late-round drift accumulates the empirical signature of Theorem~\ref{thm:bsel}. Removing $V'_k$ slows convergence; removing $F'_k$ and $St'_k$ trades terminal quality for slightly faster early progress, confirming that they are stability regularizers rather than convergence accelerators.

\begin{figure}[!t]
\centering
\includegraphics[width=\linewidth]{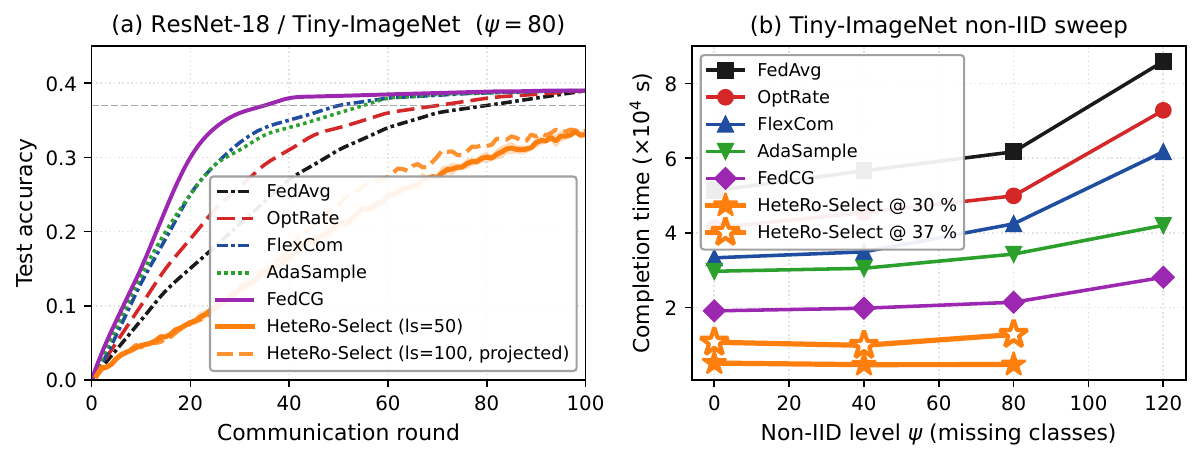}
\caption{ResNet-18 on Tiny-ImageNet (seed $42$). (a) Accuracy vs.\ round ($\psi\!=\!80$): solid = $H\!=\!50$ (hits $30\%$ at $r\!=\!79$); dashed = $H\!=\!100$ (also hits $37\%$ FedCG target). (b) Completion time to $30\%$ (filled stars) and $37\%$ (open stars) vs.\ $\psi$; baselines from~\cite{jiang2023fedcg}.}
\label{fig:ab_psi}
\end{figure}

\noindent\textbf{Operating regime ($\mu$ and $\psi$).} The $\mu$ sweep (Fig.~\ref{fig:ablations}(b)) identifies $\mu\!=\!0.1$ as the minimum regularization that makes diversity-aware selection reliable: $\mu\!=\!0.01$ reaches target two rounds earlier but drops $1.80$\,pts to final, while $\mu\!\in\!\{0,0.5\}$ fails entirely. The $\psi$ sweep shows HeteRo-Select hits the $70\%$ target for $\psi\!<\!0.6$ and fails at $\psi\!\geq\!0.6$, where the $B^2$ heterogeneity term grows beyond what score-proportional sampling can absorb; the Tiny-ImageNet missing-class sweep (Fig.~\ref{fig:ab_psi}) replicates this pattern, with the residual gap at $\psi\!=\!120$ closing when $H$ is doubled a local-compute trade-off, not a structural limit.

\begin{table}[t]
\centering
\caption{CIFAR-10 ablations ($\mu\!=\!0.1$, seed $42$ unless varied; \textbf{primary} = config used in Table~\ref{tab:resource}). $\dagger$Uniform LR outperforms score-adaptive variant; see Section~\ref{ssec:ablations}.}
\label{tab:ablations}
\resizebox{\linewidth}{!}{%
\begin{tabular}{llccccc}
\toprule
\textbf{Ablation} & \textbf{Variant} & \textbf{Peak (\%)} & \textbf{Final (\%)} & \textbf{R$\to$70\%} & \textbf{Time (s)} & \textbf{Traffic (MB)} \\
\midrule
\multirow{2}{*}{Compression}
  & Uniform & 72.22 & 72.22 & 86 & 3{,}009 & 2{,}124 \\
  & \textbf{Adaptive (primary)} & \textbf{73.03} & \textbf{73.03} & \textbf{83} & \textbf{2{,}913} & \textbf{2{,}010} \\
\midrule
\multirow{4}{*}{Proximal $\mu$}
  & $\mu\!=\!0$ & 64.36 & 64.36 & --- & --- & --- \\
  & $\mu\!=\!0.01$ & 71.92 & 70.12 & 81 & 2{,}864 & 1{,}967 \\
  & \textbf{$\mu\!=\!0.1$ (primary)} & \textbf{73.03} & \textbf{73.03} & \textbf{83} & \textbf{2{,}913} & \textbf{2{,}010} \\
  & $\mu\!=\!0.5$ & 54.57 & 53.81 & --- & --- & --- \\
\midrule
\multirow{5}{*}{Non-IID $\psi$}
  & $\psi\!=\!0$ (IID) & 73.71 & 72.61 & 77 & 2{,}655 & 1{,}933 \\
  & $\psi\!=\!0.2$ (mild) & 73.88 & 73.68 & 76 & 2{,}746 & 1{,}921 \\
  & \textbf{$\psi\!=\!0.4$ (primary)} & \textbf{73.03} & \textbf{73.03} & \textbf{83} & \textbf{2{,}913} & \textbf{2{,}010} \\
  & $\psi\!=\!0.6$ (severe) & 68.18 & 66.91 & --- & --- & --- \\
  & $\psi\!=\!0.8$ (extreme) & 59.21 & 57.02 & --- & --- & --- \\
\midrule
\multirow{3}{*}{\shortstack[l]{Score\\components}}
  & w/o $V'_k$ & 72.71 & 72.71 & 85 & 2{,}972 & 2{,}050 \\
  & w/o $D_k$ & 72.25 & 71.10 & 85 & 2{,}868 & 2{,}047 \\
  & w/o $F'_k, St'_k$ & 72.08 & 71.75 & 82 & 2{,}850 & 1{,}998 \\
\midrule
\multirow{4}{*}{\shortstack[l]{Design\\choices}}
  & Pure magnitude top-$k$ & 73.24 & 73.24 & 79 & 2{,}725 & 1{,}955 \\
  & Static $\beta\!=\!0.90$ & 72.13 & 71.97 & 83 & 2{,}865 & 2{,}005 \\
  & \textbf{Uniform LR}$^\dagger$ & \textbf{74.08} & \textbf{74.08} & \textbf{70} & \textbf{2{,}458} & \textbf{1{,}802} \\
  & Uniform aggregation & 72.28 & 72.28 & 83 & 2{,}901 & 2{,}005 \\
\bottomrule
\end{tabular}}
\end{table}

\subsection{Score Behavior and Selection Fairness}\label{ssec:score_dynamics}

\begin{figure}[t]
\centering
\includegraphics[width=\linewidth]{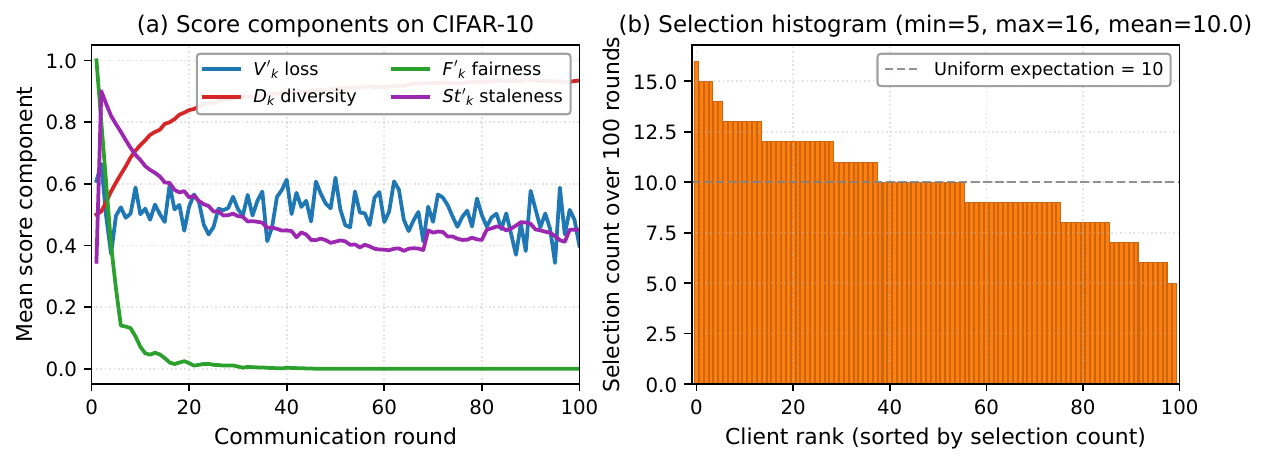}
\caption{Score-driven behavior on CIFAR-10 (seed $42$). (a)~Mean $V'_k$, $D_k$, $F'_k$, $St'_k$ over selected clients per round. (b)~Per-client selection histogram sorted descending by count; dashed line: uniform expectation $10$.}
\label{fig:mechanism}
\end{figure}

Fig.~\ref{fig:mechanism}(a) traces the four score components on CIFAR-10 over $100$ rounds. $F'_k$ dominates at cold start. Every client begins with identical selection counts, so the fairness term is the only source of differentiation and contracts within $\sim\!10$ rounds once the counts diverge. $D_k$ takes over near round $20$ and remains the dominant driver for the remainder of training, reflecting the fact that gradient directions stabilize early and the diversity signal becomes an increasingly reliable proxy for under-represented data. $V'_k$ and $St'_k$ remain in supporting ranges throughout, all four components strictly positive at every round. The dominance of $D_k$ in the middle and late phases is consistent with the ablation finding of Section~\ref{ssec:ablations}: removing $D_k$ is the only perturbation that produces a non-zero peak-to-final gap ($1.2$ pts), because without the diversity signal, the selector concentrates progressively on a narrow high-loss subset and late-round drift is no longer suppressed.

The per-client selection histogram (Fig.~\ref{fig:mechanism}(b)) is tighter than i.i.d.\ uniform sampling ($\mathrm{Bin}(100,\,0.1)$): minimum count $5$, maximum $16$, against an expectation of $10$. No client is starved; no client monopolizes, a direct consequence of $F'_k$ and $St'_k$, while $D_k$ prevents the distribution from collapsing onto the same diverse subset round after round.

\subsection{Stress Test: Inverted Informativeness--Bandwidth Coupling}\label{ssec:stress}

\begin{figure}[t]
\centering
\includegraphics[width=\linewidth]{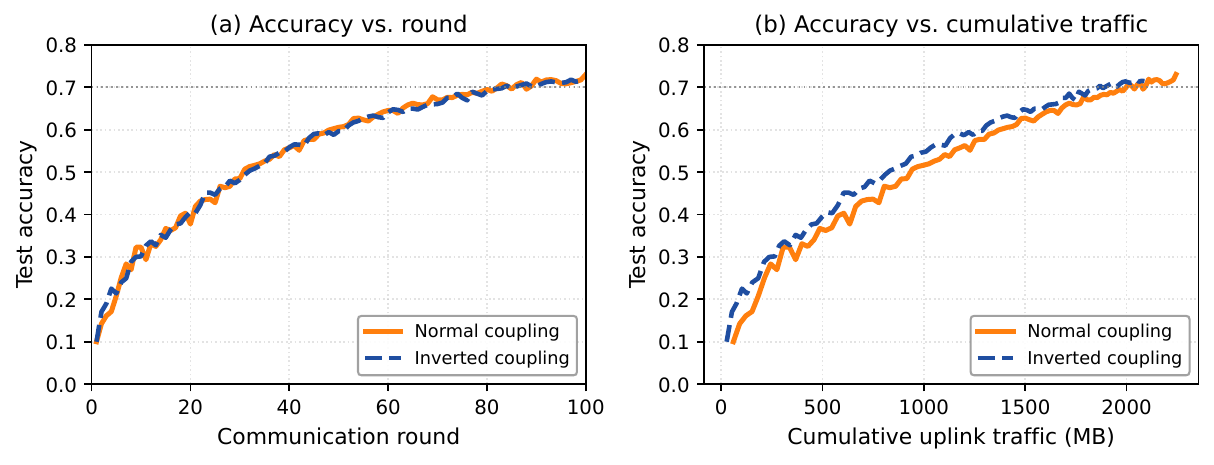}
\caption{Stress test: inverted informativeness--bandwidth coupling (CIFAR-10, $\psi\!=\!0.4$, seed $42$). (a)~Accuracy vs.\ round. (b)~Accuracy vs.\ cumulative uplink traffic.}
\label{fig:stress_curve}
\end{figure}

We assign each client's per-round uplink bandwidth inversely to its local loss, so the most informative gradients sit on the slowest links, the worst-case correlation for any informativeness-driven allocator. Under this inversion, HeteRo-Select still reaches the $70\%$ CIFAR-10 target in one extra round and with \emph{less} cumulative traffic than the matched normal-bandwidth run; the inverted curve lies below the normal curve against cumulative traffic (Fig.~\ref{fig:stress_curve}(b)) at every accuracy level above $\sim\!45\%$. The mechanism is the one Eq.~\eqref{eq:theta_k} encodes: when the bandwidth ceiling $\theta_k^{\mathrm{cap}}$ binds on an informative client, the unsent gradient mass enters the adaptive error buffer of Eq.~\eqref{eq:beta} and is transmitted in a subsequent round once bandwidth permits. A bandwidth deficit is therefore converted into a small time deficit rather than into an accuracy deficit. This is precisely the behavior the design intends: informativeness drives \emph{what} is sent, bandwidth determines \emph{when} it arrives. A symmetric run of FedCG would compound the inversion: its linear program assigns the most aggressive compression to the slowest, here most informative, clients, and FedCG carries no per-client error buffer to recover that mass, so the comparison is omitted.

\subsection{CIFAR-100 Communication Efficiency}\label{ssec:cifar100}

On CIFAR-100, the relevant claim is communication efficiency at every accuracy level actually attained, not terminal accuracy. Under the same $T\!=\!100$ budget, HeteRo-Select reaches any given test accuracy using $45.2\%$ of FedCG's simulated time and $59.6\%$ of its simulated traffic (Table~\ref{tab:resource}), with seed-invariant cumulative resource usage across three seeds. The $54\%$ FedCG target is not crossed within $100$ rounds at the $40$-class-missing skew level: the label distribution is extreme enough that the $B^2$ heterogeneity term grows beyond what the score-proportional selector can absorb in a fixed budget, the same regime in which the $\psi$ sweep of Section~\ref{ssec:ablations} shows the CIFAR-10 target failing at $\psi\!\geq\!0.6$. This is a compute-budget issue rather than a per-round design issue: the same per-round efficiency advantage that holds at every lower accuracy level implies that a longer run would continue to close the gap with FedCG at the same traffic rate.

\subsection{Cross-Scale Generalization}\label{ssec:scale}

To verify the same configuration is not over-tuned to the AlexNet/ResNet9 pair, we run MNIST and TinyImageNet with no hyperparameter retuning under the identical FedCG-matched protocol (seed $42$). Table~\ref{tab:scale} reports the per-dataset summary.

\begin{table}[t]
\centering
\caption{Cross-scale generalization ($\mu\!=\!0.1$, seed $42$; same config as Table~\ref{tab:resource}). Three successful benchmarks span $\sim\!3{,}400\times$ in model size.}
\label{tab:scale}
\resizebox{\linewidth}{!}{%
\begin{tabular}{lccccc}
\toprule
\textbf{Dataset / Model} & \textbf{Params} & \textbf{Peak (\%)} & \textbf{R$\to$target} & \textbf{Time (s)} & \textbf{Traffic (MB)} \\
\midrule
MNIST / LogReg           & $7{,}850$ & 91.94 & 5   & 118.3   & 0.61    \\
CIFAR-10 / AlexNet       & $2.78$M   & 73.03 & 83  & 2{,}913 & 2{,}010 \\
CIFAR-100 / ResNet9      & $6.62$M   & 49.44 & --- & ---     & ---     \\
TinyImageNet / ResNet-18 & $11.27$M  & 33.53 & 79  & 4{,}622 & 7{,}061 \\
\bottomrule
\end{tabular}}
\end{table}

Table~\ref{tab:scale} shows that the same configuration hits the per-dataset target on three of four benchmarks across a $\sim\!3{,}400\times$ range in model size, with TinyImageNet/ResNet-18 stable in late rounds. The only un-hit target is CIFAR-100, consistent with the budget-at-extreme-heterogeneity reading of Section~\ref{ssec:cifar100}, since the deeper TinyImageNet model reaches its proportionally calibrated target in the same $100$ rounds.

\section{Conclusion}\label{sec:conclusion}

We presented HeteRo-Select, which treats informativeness, rather than bandwidth, as the primary driver of client participation, tying client selection, gradient-compression ratio, and server-side aggregation weight to a single composite score with bandwidth retained only as a hard ceiling. Under a FedCG-matched protocol, the design improves time- and traffic-to-target on CIFAR-10, generalizes from a $7{,}850$-parameter LR to an $11.27$M-parameter ResNet-18 without retuning, and remains robust when informativeness and bandwidth are deliberately anti-correlated. An ablation further reveals that score-adaptive local learning-rate over-weighting of informativeness in the local optimizer; a globally decaying schedule produces strictly better results, identifying the communication layer as the effective scope of the informativeness signal. Two limitations remain: at the most extreme CIFAR-100 label skew, $100$ rounds are insufficient to reach the FedCG target accuracy (communication efficiency at every lower accuracy level is still improved), and the Hutchinson curvature probe adds a single diagonal estimate per selected client per round, layer-budgeted to $Q\!=\!3$, a single-digit-percentage overhead on the dominant $H\!=\!50$ local FedProx steps, and one that does not recover its cost in accuracy at AlexNet scale. Future work will extend the analysis to heavy-tailed gradient noise, integrate quantization with score-proportional sparsification, and evaluate on a real prototype with measured rather than simulated bandwidth.

\bibliographystyle{IEEEtran}
\bibliography{main}

\vfill

\end{document}